\documentclass[conference]{IEEEtran}
\IEEEoverridecommandlockouts

\usepackage{cite}
\usepackage{amsmath,amssymb,amsfonts}
\usepackage{algorithmic}
\usepackage{graphicx}
\usepackage{booktabs}
\usepackage{multirow}
\usepackage{textcomp}
\usepackage{xcolor}
\usepackage{comment}
\usepackage[table]{xcolor}
\usepackage{booktabs}
\usepackage{subcaption}
\usepackage{url}

\definecolor{usable}{HTML}{91BFDB}
\definecolor{weak}{HTML}{FFFFBF}   
\definecolor{degen}{HTML}{FC8D59}

\def\BibTeX{{\rm B\kern-.05em{\sc i\kern-.025em b}\kern-.08em
    T\kern-.1667em\lower.7ex\hbox{E}\kern-.125emX}}
\begin{document}

\title{Bias Analysis of L2 Speaking Assessment Systems Using Concept Activation Vectors}


\author{\IEEEauthorblockN{Arya Labroo}
\IEEEauthorblockA{\textit{University of Cambridge} \\
al2135@cam.ac.uk}
\and
\IEEEauthorblockN{Mengjie Qian}
\IEEEauthorblockA{\textit{University of Cambridge} \\
mq227@cam.ac.uk}
\and
\IEEEauthorblockN{Kate Knill}
\IEEEauthorblockA{\textit{University of Cambridge} \\
kmk1001@cam.ac.uk}
}

\maketitle

\begin{abstract}

Automatic speaking assessment systems are increasingly deployed in high-stakes settings to mark second language (L2) learners' speaking tests, making it critical to show that their scores depend on speaking proficiency rather than irrelevant speaker attributes such as first language (L1) or age. Transformer-based foundation models have improved the accuracy of these L2 speaking graders, but their black-box representations make fairness and interpretability analysis more difficult. Building on prior work that used Concept Activation Vectors (CAVs) to detect bias towards unwanted attributes (`concepts') in feature-based graders, we extend CAV-based analysis to two neural speaking assessment systems: a text-based BERT grader and a speech-and-text multimodal grader based on Whisper. CAVs represent human-interpretable concepts as directions in a model's activation space, allowing us to distinguish between whether a concept is encoded in a model's internal representations and whether it influences the predicted score, the latter quantified using a gradient-based sensitivity metric. Since CAVs rely on linear separability, which is less likely in complex neural embedding spaces, we also investigate whether sparse autoencoders (SAEs) provide cleaner concept directions by learning CAVs in a sparse latent space and mapping them back to activation space. Our analysis shows that concept recoverability depends strongly on the representation and architecture being probed, rather than on the concept alone. Sensitivity to concepts is also architecture-dependent. SAEs make concepts more linearly recoverable, but attenuate the original activation-space sensitivity, especially in low-dimensional layers. These findings highlight the need to distinguish concept recoverability from concept influence when auditing bias in speaking assessment systems.

\end{abstract}


\section{Introduction}
\label{sec:intro}

Second language (L2) automatic speaking assessment systems estimate a candidate's proficiency level from their spoken responses to one or more questions. The goal of these systems is to match the scores a human examiner would award to the candidate.
Transformer foundation model based versions of these systems have achieved state-of-the-art performance which meet the scoring accuracy requirements e.g.~\cite{raina20_interspeech, banno2022slt, banno2023assessment, mcknight23_slate, ma2025b_interspeech, lin25_slate, phan25_slate}. These systems take audio and/or text (ASR transcripts) as input, convert the input into an embedding representation which is then passed to a regression head that predicts the candidate's test score. Models such as BERT~\cite{devlin2019bert}, wav2vec2.0~\cite{baevski2020wav2vec2} and Whisper~\cite{radford2022whisper} have been used to encode the input from the candidate.

As these foundation model based automatic speaking assessment systems are increasingly deployed in high-stakes settings such as immigration and employment certification, it becomes critical that their scores reflect speaking ability rather than irrelevant attributes of the speaker. A fair grader should rely on evidence of ability such as fluency, grammatical control, and content, and remain insensitive to attributes such as first language (L1) or age. These attributes are nonetheless richly present in the data and can enter a grader either from the large, under-documented corpora used to pre-train its foundation model, or from imbalance in the task-specific fine-tuning data, where an over-represented group at a given proficiency can become a shortcut for predicting the score.

Such biases have been documented across a wide range of machine learning applications, including demographic accuracy disparities in commercial gender classification and face analysis systems \cite{buolamwini2018gendershades} and gender stereotypes encoded in word embeddings \cite{bolukbasi2016debiasing}. 
However, detecting bias in modern neural models is more challenging, because aggregate performance metrics do not reveal which internal representations drive a model's predictions, and these representations are often highly distributed and entangled.
A line of interpretability work has shown, however, that human-interpretable information can be recovered from a network's internal activations: linear classifier probes can decode properties from intermediate layers \cite{alain2017probes}, and individual representational directions can be aligned with semantic concepts \cite{bau2017networkdissection}. 
Building on these ideas, the TCAV framework \cite{kim2018tcav} introduces Concept Activation Vectors (CAVs), which represent a human-interpretable concept as a direction in the activation space learned by a linear classifier to separate activations of examples that exhibit the concept from those that do not. The influence of a concept is then quantified by measuring the sensitivity of the model's prediction to that direction using gradients. 
Applied to image classification, TCAV has been used both to explain model predictions and to uncover biases, for example by associating the concept `female' with the class `apron'. Crucially, CAVs distinguish between whether a concept is encoded in a model's representations and whether it actually influences the model's prediction.

Wei et al. \cite{wei2021cavsla} adapted CAVs from classification to the regression setting of automatic speaking assessment and used them to detect bias in a feature-based grader, i.e. one based on `hand-crafted' and human interpretable features. They validated the method by recovering a deliberately injected L1 bias and showing that concepts flagged on training data corresponded to bias on expert-scored evaluation data. 
This work extend their analysis to Transformer-based graders, showing that the linear recoverability of a concept is not a reliable indicator of its influence on the predicted score. Building on this observation, we investigate how concept recoverability and influence vary across the layers of modern Transformer foundation models, and whether learning CAVs in a sparse latent representation improves concept-based bias analysis. This question is motivated by a key assumption of CAVs that a concept can be represented by a single linear direction, an assumption that becomes increasingly restrictive as Transformer representations become higher-dimensional and more entangled.
Sparse autoencoders (SAEs), which learn a sparse, higher-dimensional latent space in which features may be more disentangled, have recently been shown to recover more interpretable features from Transformer activations than the raw neurons \cite{bricken2023monosemanticity, gao2024scaling}.
This raises a natural question: does learning CAVs in an SAE latent space improve concept-based bias analysis?

This work aims to address these challenges with the following contributions: i) we extend CAV-based bias analysis to Transformer-based automatic speaking assessment systems; ii) we compare representation-level bias across text-only (BERT) \cite{devlin2019bert} and speech-text (Whisper) graders~\cite{benedict2024thesis}, iii) we investigate sparse autoencoders as a complimentary representation for concept-based bias analysis.

In the rest of this paper, Sections \ref{sec:cav} and \ref{sec:sae} describe the CAV and SAE methods used to assess and probe a grader's sensitivity to a concept. Section \ref{sec:setup} describes the datasets, graders, and concepts, and Section \ref{sec:experiments} presents the experiments, comparing concept recoverability and concept influence across both architectures and both probing methods.

\section{Concept Activation Vectors}
\label{sec:cav}
Concept Activation Vectors (CAVs)~\cite{kim2018tcav} provide a representation-level method for analysing whether a neural network encodes human-interpretable concepts and whether those concepts influence its predictions. In the context of spoken language assessment, this enables us to distinguish whether attributes such as L1, gender, or proficiency are merely represented by the model or contribute to the predicted score. A trained grader with parameters $\theta$ maps an input response $\mathbf{x}^{(i)}$ for speaker $i$ to a predicted score
\begin{equation}
    \hat{y}^{(i)} = F(\mathbf{x}^{(i)}; \theta).
    \label{eq:cav_model}
\end{equation}
For CAV analysis the model can be split at a chosen internal layer, mapping the input to the activation vector $\mathbf{h}^{(i)} \in \mathbb{R}^d$ at that layer, and then that activation to the output,
\begin{equation}
    \mathbf{h}^{(i)} = F_h(\mathbf{x}^{(i)};\theta); \qquad
    \hat{y}^{(i)} = F_y(\mathbf{h}^{(i)};\theta).
    \label{eq:cav_split}
\end{equation}
The key idea is that the hidden activations form a space in which properties of the input may be represented geometrically: if examples corresponding to a concept $c$, occupy a different region from examples not corresponding to it, then a direction in activation space captures the difference between the two groups, and this direction is the CAV. 

To construct it, each example is given a binary concept label
$t^{(i)} = +1$ if it belongs to concept $c$ and $t^{(i)} = -1$ otherwise. Then a linear classifier is trained to separate the positive and negative activation vectors by minimising the hinge loss
\begin{equation}
    \mathcal{L}(\mathbf{d}^{(c)}, b)
    =
    \sum_{i=1}^{N}
    \max\!\left\{
        0,\,
        1 - t^{(i)}\!\left((\mathbf{d}^{(c)})^\top \mathbf{h}^{(i)} + b\right)
    \right\},
    \label{eq:cav_hinge}
\end{equation}
with an $L_2$ penalty on $\mathbf{d}^{(c)}$ and balanced class weighting to account for concept imbalance. The vector $\mathbf{d}^{(c)} \in \mathbb{R}^d$, normal to the decision boundary, is the CAV for concept $c$, and its direction points towards increasing presence of the concept. If the classifier cannot separate the classes better than chance, the concept is not linearly recoverable at that layer and the CAV is uninformative. 
Conversely, high classification accuracy indicates only that the concept is encoded in the representation, not that it influences the predicted score; influence is assessed using the sensitivity metric described below.

The original TCAV~\cite{kim2018tcav} method was developed for classification. In spoken language assessment the output is a scalar proficiency score, so the question becomes whether movement in the concept direction changes the predicted score. Following Wei et al ~\cite{wei2021cavsla}, which adapted CAVs to regression, we analyse the alignment between the concept direction and the gradient of the predicted score with respect to the activation,
\begin{equation}
    \nabla_{\mathbf{h}} F_y(\mathbf{h}^{(i)};\theta)
    =
    \left.\frac{\partial F_y(\mathbf{h};\theta)}{\partial \mathbf{h}}\right|_{\mathbf{h}=\mathbf{h}^{(i)}}.
    \label{eq:cav_grad}
\end{equation}
In order for CAV magnitudes to be comparable across layers and models, the direction is scaled by the mean activation norm,
\begin{equation}
    \Delta\mathbf{h}^{(c)}
    =
    \left(\frac{1}{N}\sum_{i=1}^{N}\|\mathbf{h}^{(i)}\|_2\right)
    \frac{\mathbf{d}^{(c)}}{\|\mathbf{d}^{(c)}\|_2}.
    \label{eq:cav_scaled}
\end{equation}
Comparing directions with the cosine distance $\cos(\mathbf{a},\mathbf{b}) = 1 - \mathbf{a}^\top\mathbf{b} / (\|\mathbf{a}\|_2\|\mathbf{b}\|_2)$, which is $0$ for aligned, $1$ for orthogonal, and $2$ for opposite directions, the main sensitivity metric is the mean per-speaker gradient cosine distance,
\begin{equation}
    B^{(c)}_{\mathrm{gr}}
    =
    \frac{1}{N}\sum_{i=1}^{N}
    \cos\!\left(
        \Delta\mathbf{h}^{(c)},
        \nabla_{\mathbf{h}} F_y(\mathbf{h}^{(i)};\theta)
    \right).
    \label{eq:cav_bgr}
\end{equation}
A value of $B^{(c)}_{\mathrm{gr}}$ near $1$ indicates that the concept direction is close to orthogonal to the score-increasing direction, so the score is insensitive to the concept; a value below $1$ indicates that increasing the concept tends to increase the predicted score, while a value above $1$ indicates that it tends to decrease it. Potential bias is therefore indicated only when a concept that should be irrelevant to scoring is both recoverable, by classifier accuracy, and also aligned with the scoring direction, by $B^{(c)}_{\mathrm{gr}}$.

\section{Sparse Autoencoders}
\label{sec:sae}

A CAV assumes a concept can be captured by a single linear direction. While this is often reasonable for smaller models, it weakens as models scale, since large speech and language models tend to represent many factors in overlapping, non-orthogonal directions, known as superposition. This creates two problems: a concept may be encoded but not linearly separable; and even when a linear classifier does separate the examples, the resulting direction may mix the intended concept with other correlated factors. 
Sparse autoencoders (SAEs) learn sparse, higher-dimensional representations that have been shown to recover more interpretable features from Transformer activations~\cite{bricken2023monosemanticity, gao2024scaling}. We therefore investigate whether learning CAVs in the SAE latent space yields cleaner concept directions for bias analysis.


We adopt the TopK SAE \cite{gao2024scaling}, which enforces sparsity explicitly by retaining only the $K$ largest latent activations for each input. An SAE has two parts: an encoder that maps an activation to a higher-dimensional latent vector, and a linear decoder that reconstructs the activation from it. Given an activation vector $\mathbf{h} \in \mathbb{R}^d$ at the layer under analysis, the encoder produces a latent vector
$\mathbf{z} \in \mathbb{R}^m$ (with $m > d$) by keeping only its $K$ largest entries, and the decoder reconstructs $\hat{\mathbf{h}}$,
\begin{equation}
    \mathbf{z}=\mathrm{TopK}\!\left(\mathrm{ReLU}(\mathbf{W}_E\mathbf{h}+\mathbf{b}_E)\right),\qquad
    \hat{\mathbf{h}}=\mathbf{W}_D\mathbf{z}+\mathbf{b}_D,
    \label{eq:topk}
\end{equation}
where $\mathbf{W}_E\in\mathbb{R}^{m\times d}$ and $\mathbf{b}_E$ are the encoder weights and bias, $\mathbf{W}_D\in\mathbb{R}^{d\times m}$ and
$\mathbf{b}_D$ the decoder weights and bias, $d$ is the activation dimension, and $m$ is the latent dimension. The $\mathrm{TopK}$ operation zeroes all but the $K$ largest entries, so each latent vector has at most $K$ active units.
The SAE is trained on activations extracted from the frozen grader by minimising the reconstruction error $\|\mathbf{h} \hat{\mathbf{h}}\|_2^2$; the grader is not trained jointly.

Once trained, the SAE provides an alternative representation for learning concept directions. Rather than training the concept classifier directly on the dense activations $\mathbf{h}$, each activation is first encoded as a sparse latent vector $\mathbf{z}$, and the CAV is learned in the latent space, producing a latent-space CAV $\mathbf{d}^{(c)}_z$.

The linear decoder gives the SAE a useful geometric interpretation. Writing the reconstruction as 
    $\hat{\mathbf{h}} = \sum_{j=1}^{m} z_j \mathbf{w}_j + \mathbf{b}_D$
each decoder column $\mathbf{w}_j \in \mathbb{R}^d$ is a direction in the original activation space, so a linear direction in the latent space corresponds to a weighted combination of decoder directions in the activation space. This makes it possible to learn a CAV in the sparse latent space and map it back into activation space, where it can be evaluated with the same gradient metric as the baseline CAV.
Specifically, each activation is first encoded, and a linear classifier is trained to separate the concept classes in the latent space, giving a latent space CAV $\mathbf{d}_z^{(c)} \in \mathbb{R}^m$. 

Since the decoder is linear, 
the corresponding activation-space CAV is obtained by
\begin{equation}
    \mathbf{d}^{(c)}_{h,\mathrm{SAE}}
    =
    \frac{\mathbf{W}_D\mathbf{d}_z^{(c)}}{\|\mathbf{W}_D\mathbf{d}_z^{(c)}\|_2},
    \label{eq:sae_cav_map}
\end{equation}
The mapped CAV is evaluated using the same gradient-based sensitivity metric $B_{\mathrm{gr}}^{(c)}$ as the standard CAV. 
The standard CAV and SAE-CAV approaches therefore differ only in whether the concept direction is learned from the dense activations $\mathbf{h}$ or from the sparse vectors $\mathbf{z}$. The mapped direction lies in the span of the decoder columns, and so is constrained by what the SAE has learned to reconstruct - if the SAE does not reconstruct the part of activation space relevant to a concept well, the mapped CAV may be unreliable. The latent space intercept also does not transfer cleanly, since the encoder is non-linear, so the mapped quantity is interpreted as a direction only.

The hyperparameters for the SAE are chosen based on four complementary criteria: reconstruction error, sparsity, dead-feature fraction, and decoder coherence. Reconstruction error, measured by the mean squared error between $\mathbf{h}$ and $\hat{\mathbf{h}}$, indicates how well the sparse code preserves the original activations. Sparsity, the fraction of latent units inactive (zero) on average, controls whether individual latent units can be interpreted as distinct features. For a TopK SAE, this is fixed at $1 - K/m$. The dead-feature fraction is the proportion of latent units that never activate on the data, and measures wasted capacity. Finally, decoder coherence, the mean absolute cosine similarity between decoder columns, captures redundancy: a value near zero indicates the feature directions are close to orthogonal rather than overlapping. These trade off against one another, and so are considered jointly. 

\section{Experimental Setup}
\label{sec:setup}

\subsection{Datasets}
\label{subsec:data}

The primary dataset is the Speaking component of the Business Language Testing Service (BULATS) examination~\cite{chambers2011bulats}, a multi-level monologic computer assisted test of business-focused L2 English \cite{wei2021cavsla}.
Responses are scored by trained human examiners on a scale from $0$ to $6$ and mapped to CEFR~\cite{cefr2001}  levels pre-A1 to C2. Our analysis focuses on Part~1, in which the candidate answers six short questions about themselves and their background, work, or future plans, producing short, semi-spontaneous personal responses well suited to per-response representation analysis. The data comprises $1657$ training, $188$ development, and $465$ evaluation speakers. Importantly, this dataset includes rich speaker metadata, including gender, age, and L1, enabling analysis of multiple potential sources of bias.

As a secondary corpus we use the Speak \& Improve (S\&I) 2025 Corpus \cite{sicorpus25,knill2025speakimprove}, a publicly available dataset of L2 learner English with a closely related computer-delivered, monologic Part~1 format, comprising $3068$ training, $438$ development, and $300$ evaluation speakers. 
is used to assess whether findings on BULATS generalise to an independently collected corpus, while its public availability supports reproducibility.
Its main limitation for the purposes of this work is the absence of demographic metadata. 
To enable demographic analysis, we infer speaker gender using an ECAPA-TDNN~\cite{desplanques20_interspeech} voice gender classifier \cite{huh_voice_gender_classifier_github}. Other demographic concepts are not considered, as they cannot be reliably inferred using existing external models.

\subsection{Grader Models}
\label{subsec:models}

We analyse two graders that differ in input modality and model architecture. The BERT grader takes the response transcript as input and outputs a holistic proficiency score~\cite{qian2024speak, qian25_slate}. The transcriptions are obtained from an automatic speech recognition (ASR) system, here, the out-of-the-box Whisper small model~\cite{radford2022whisper} is used. The BERT grader produces token embeddings, which are pooled into a fixed-dimensional response representation by four parallel self-attention heads and passed through a feed-forward regression head with two hidden layers and a linear output. Analysis is
performed at two points in this head: the post-ReLU output of the first hidden layer ($d = 600$), denoted \texttt{Layer 1}, and the post-ReLU output of the second ($d = 20$), denoted \texttt{Layer 2}. The grader is fine-tuned end-to-end with an MSE objective.

The Whisper grader is a `Whisper-Fuse' model built on a pre-trained encoder-decoder \cite{radford2022whisper}, which additionally accepts the audio signal. Each Part~1 question response is a segment: its audio is converted to a log-mel spectrogram and encoded into latent audio embeddings, and the decoder cross-attends to these while running over the segment's
transcript tokens, fusing acoustic and lexical information. Attention pooling collapses each segment to a fixed vector, a second stage of attention pooling combines the segments, and the result is passed to a head of two linear layers with a ReLU between them. Analysis is performed at the $2048$-dimensional concatenated input to the head, \texttt{dense.in}, and the
$64$-dimensional post-ReLU representation, \texttt{act.out}. This grader was not trained as part of this work, instead a provided checkpoint is analysed~\cite{benedict2024thesis}.

\subsection{Concepts}
\label{subsec:concepts}

A concept is defined by a binary labelling of speakers. \emph{Proficiency} concepts are derived by thresholding the CEFR scale, e.g. $\ge\!A2$ ($\ge\!2.0$) or $\ge\!B2$ ($\ge\!4.0$). These are expected to align with the score and so verify that the pipeline detects influence when it should be
present. \emph{Demographic} and \emph{L1} concepts are the concepts of primary interest, since the grader should be insensitive to them. BULATS provides gender (the positive class is female), age, and several first languages, while S\&I provides only the inferred gender concept described above. Table~\ref{tab:concepts} summarises the mean score of each BULATS concept's positive and negative groups. A non-zero gap $\Delta$ means the concept correlates in some way with proficiency in the data, against which any measured sensitivity must be read.

\begin{table}[t]
\centering
\small
\setlength{\tabcolsep}{5pt}
\caption{Mean BULATS score by concept. $n_{+}$/$n_{-}$ are the
positive/negative class sizes, $\bar{s}_{+}$/$\bar{s}_{-}$ their mean scores,
and $\Delta = \bar{s}_{+} - \bar{s}_{-}$. The positive class for Gender is
female.}
\label{tab:concepts}
\begin{tabular}{lrrrrr}
\toprule
Concept & $n_{+}$ & $\bar{s}_{+}$ & $n_{-}$ & $\bar{s}_{-}$ & $\Delta$ \\
\midrule
\multicolumn{6}{l}{\textit{Proficiency (CEFR threshold)}} \\
$\ge\!A2$      & 1579 & 3.82 & 78   & 1.31 & 2.51 \\
$\ge\!B1$      & 1344 & 4.09 & 313  & 2.06 & 2.03 \\
$\ge\!B2$      & 862  & 4.56 & 795  & 2.79 & 1.77 \\
\addlinespace
\multicolumn{6}{l}{\textit{Demographic}} \\
Gender         & 643  & 3.69 & 613  & 3.69 & $-0.00$ \\
Age $\le\!30$  & 907  & 3.63 & 743  & 3.80 & $-0.17$ \\
\addlinespace
\multicolumn{6}{l}{\textit{First language (L1)}} \\
Dutch          & 135  & 4.46 & 1522 & 3.64 & $0.82$ \\
Italian        & 111  & 4.03 & 1546 & 3.68 & $0.35$ \\
Polish         & 167  & 4.00 & 1490 & 3.67 & $0.32$ \\
Japanese       & 97   & 3.41 & 1560 & 3.72 & $-0.31$ \\
Thai           & 105  & 3.38 & 1552 & 3.73 & $-0.35$ \\
Gujarati       & 207  & 3.31 & 1450 & 3.76 & $-0.45$ \\
\bottomrule
\end{tabular}
\end{table}

For each concept and layer, activations and per-speaker score gradients are captured at the four analysis points above, and a CAV is extracted with the regularised hinge classifier of Section~\ref{sec:cav}, using balanced class weighting given the imbalance of the L1 and proficiency concepts. For each layer, a TopK SAE is swept over the latent size $m$ and sparsity $k$ and the configuration is selected to balance reconstruction error, sparsity, dead-feature fraction, and decoder coherence on the development split.

\section{Experiments}
\label{sec:experiments}

\subsection{Grader Performance}
\label{subsec:grader_perf}
Bias analysis is meaningful only if the underlying speaking assessment model provides a reliable estimate of proficiency. We therefore first evaluate the grading performance of the BERT and Whisper graders (Section~\ref{subsec:models}). Table~\ref{tab:grader_perf} reports both graders on the BULATS and S\&I evaluation sets. Pearson Correlation Coefficient (PCC) values lie between $0.68$ and $0.79$, and a large majority of responses are placed within a full grade band of the human score, implying that all four (model, dataset) pairs are suitable for subsequent analysis. The Whisper grader's lower Root Mean Square Error (RMSE) on S\&I is expected, since the grader was fine-tuned on S\&I.

\begin{table}[t]
\centering
\small
\caption{Grader performance on the BULATS and S\&I evaluation sets. $\le\!0.5$ and $\le\!1.0$ are the percentage of responses scored within $0.5$ and $1.0$ of the human score. All metrics are after calibration.}
\label{tab:grader_perf}
\begin{tabular}{llrrrr}
\toprule
Corpus & Grader & RMSE & PCC & $\le\!0.5$ & $\le\!1.0$ \\
\midrule
\multirow{2}{*}{BULATS}
 & BERT    & 0.817 & 0.756 & 45.0 & 78.2 \\
 & Whisper & 0.781 & 0.786 & 44.7 & 80.4 \\
\addlinespace
\multirow{2}{*}{S\&I}
 & BERT    & 0.557 & 0.684 & 62.3 & 93.3 \\
 & Whisper & 0.546 & 0.749 & 59.0 & 95.7 \\
\bottomrule
\end{tabular}
\end{table}

\subsection{Concept Recoverability}
\label{subsec:recoverability}

Before measuring influence, it is first necessary to determine whether each concept is linearly recoverable. Table~\ref{tab:recoverability} reports the balanced accuracy, the mean of the positive and negative concept class accuracies, of the CAV classifier on BULATS. This has a chance baseline of 50\% regardless of class prevalence. A cell is marked \emph{degenerate} (orange) when the smaller of the two per-class accuracies falls below $10\%$, meaning that the CAV has collapsed to predicting one class for every speaker and identifies the class-prior direction rather than a concept direction. A cell is marked \emph{weakly usable} (yellow) when both classes are predicted but balanced accuracy is below $60\%$, and finally \emph{usable} (blue) otherwise. We exclude degenerate cells from the bias analysis.

Whisper \texttt{dense.in} is the only layer where every concept is linearly recoverable. 
The two BERT layers and the Whisper post-ReLU output have more than half L1 concepts degenerate. The proficiency concepts are usable everywhere, as expected as they are what the graders are trained to capture. 
All four layers share the same class imbalance, yet recoverability differs substantially, i.e. only \texttt{dense.in} separates the demographic and L1 concepts. The failure of the other layers is therefore a property of the representations they learn, not of the imbalance. 
Subsequent analyses focus on Dutch and Gujarati, the only L1 concepts recoverable across both BERT layers, enabling consistent comparison across models.

\begin{table}[t]
\centering
\small
\setlength{\tabcolsep}{4pt}
\caption{BULATS CAV quality: balanced accuracy (\%) of the linear classifier on the training set. Blue = usable, yellow = weakly usable, orange = degenerate. `Pos \%' is the positive-class prevalence.}
\label{tab:recoverability}
\begin{tabular}{lr cc cc}
\toprule
& & \multicolumn{2}{c}{BERT} & \multicolumn{2}{c}{Whisper} \\
\cmidrule(lr){3-4}\cmidrule(lr){5-6}
Concept & Pos \% & Layer 1 & Layer 2 & dense.in & act.out \\
\midrule
\multicolumn{6}{l}{\textit{Proficiency}} \\
$\ge\!A2$ & 95.3 & \cellcolor{usable}88.1 & \cellcolor{usable}87.4 & \cellcolor{usable}91.8 & \cellcolor{usable}88.8 \\
$\ge\!B1$ & 81.1 & \cellcolor{usable}83.7 & \cellcolor{usable}84.0 & \cellcolor{usable}85.5 & \cellcolor{usable}82.0 \\
$\ge\!B2$ & 52.0 & \cellcolor{usable}79.7 & \cellcolor{usable}79.6 & \cellcolor{usable}82.4 & \cellcolor{usable}78.5 \\
\addlinespace
\multicolumn{6}{l}{\textit{Demographic}} \\
Gender    & 51.2 & \cellcolor{weak}57.4  & \cellcolor{weak}51.8  & \cellcolor{usable}85.0 & \cellcolor{usable}61.8 \\
Age $\le\!30$ & 55.0 & \cellcolor{usable}60.1 & \cellcolor{weak}54.3 & \cellcolor{usable}75.3 & \cellcolor{weak}59.2 \\
\addlinespace
\multicolumn{6}{l}{\textit{First language (L1)}} \\
Arabic    & 10.0 & \cellcolor{degen}50.8 & \cellcolor{degen}50.0 & \cellcolor{usable}80.7 & \cellcolor{degen}50.0 \\
Dutch     &  8.1 & \cellcolor{usable}66.2 & \cellcolor{usable}65.6 & \cellcolor{usable}88.5 & \cellcolor{usable}70.5 \\
French    &  7.3 & \cellcolor{weak}57.6  & \cellcolor{weak}57.6  & \cellcolor{usable}72.0 & \cellcolor{weak}54.5 \\
Gujarati  & 12.5 & \cellcolor{weak}59.2  & \cellcolor{usable}66.4 & \cellcolor{usable}88.9 & \cellcolor{usable}70.7 \\
Italian   &  6.7 & \cellcolor{degen}54.0 & \cellcolor{degen}53.2 & \cellcolor{usable}64.9 & \cellcolor{degen}50.0 \\
Japanese  &  5.9 & \cellcolor{degen}50.0 & \cellcolor{degen}50.0 & \cellcolor{usable}73.6 & \cellcolor{degen}50.0 \\
Polish    & 10.1 & \cellcolor{degen}51.7 & \cellcolor{degen}50.0 & \cellcolor{usable}81.8 & \cellcolor{degen}50.4 \\
Thai      &  6.3 & \cellcolor{degen}50.0 & \cellcolor{degen}50.0 & \cellcolor{usable}81.5 & \cellcolor{degen}50.0 \\
\bottomrule
\end{tabular}
\end{table}

\subsection{Bias Detection in Activation Space}
\label{subsec:bias_act}

Here we examine whether recoverable concepts influence the predicted scores. Specifically, the gradient-based sensitivity metric $B_{\mathrm{gr}}$ is applied to each (model, layer, concept) whose CAV is usable, with the activation-space columns of Table~\ref{tab:bgr} reporting the results. Since sensitivity does not imply unfairness when a concept correlates with true proficiency, each value is read against the score gaps in Table~\ref{tab:concepts}.

\begin{figure}[t]
    \centering
    \begin{subfigure}{0.473\linewidth}
        \centering
        \includegraphics[width=0.99\linewidth]{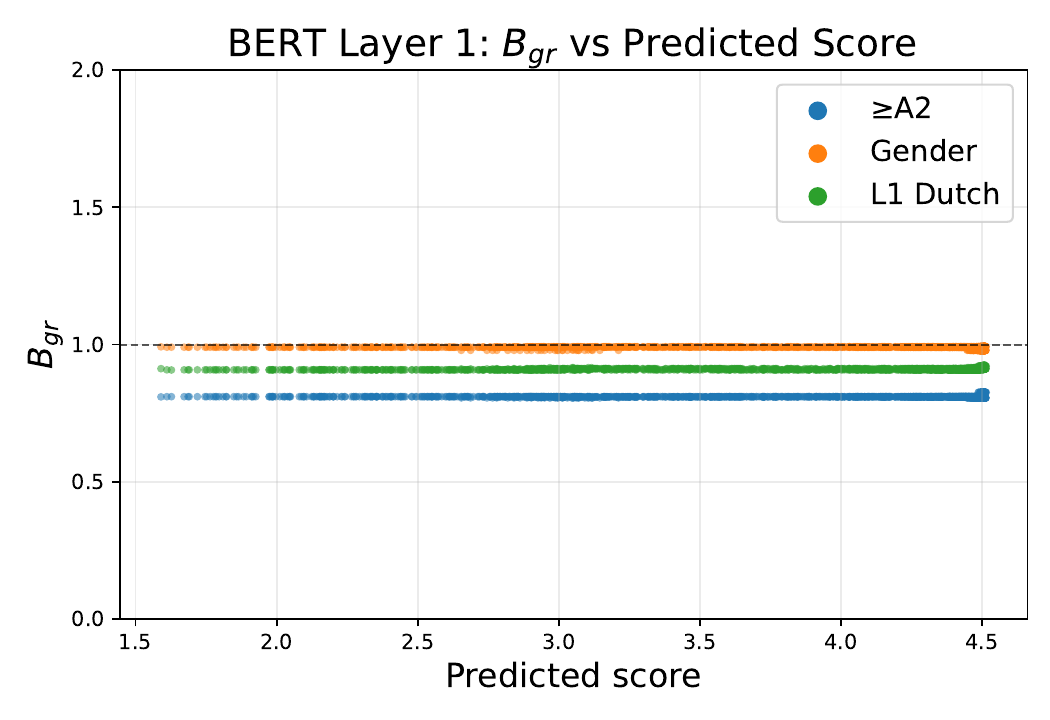}
    \end{subfigure}
    \hfill
    \begin{subfigure}{0.473\linewidth}
        \centering
        \includegraphics[width=0.99\linewidth]{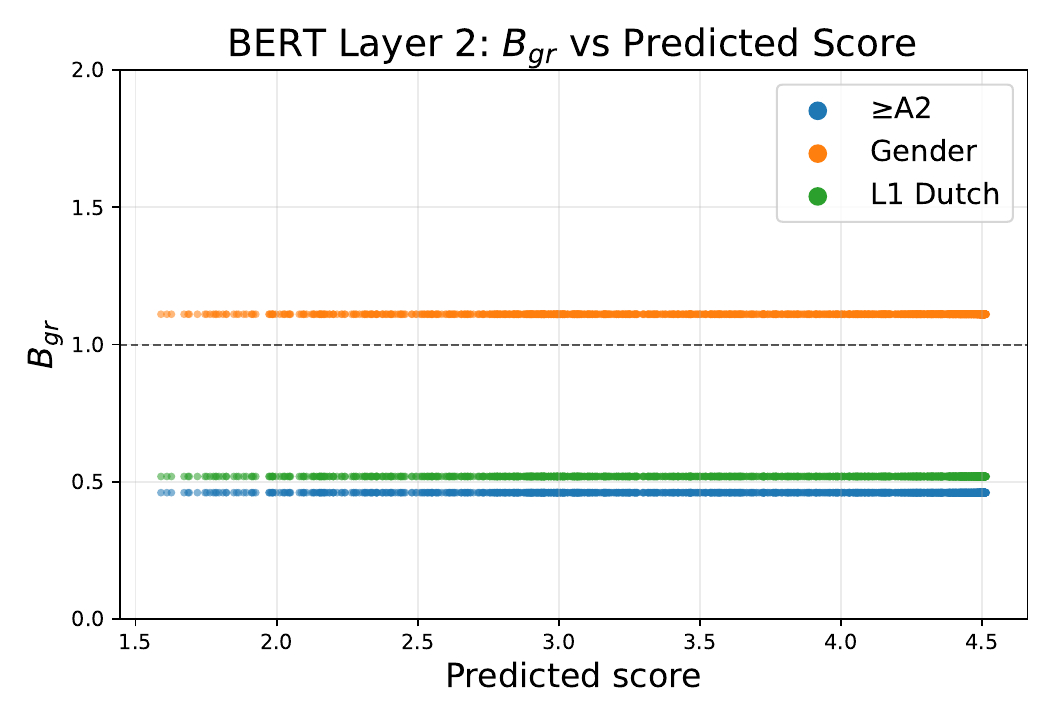}
    \end{subfigure}
    \caption{Per-speaker $B_{\mathrm{gr}}$ for the BERT grader at Layer 1 (left) and Layer 2 (right), for the $\ge\!A2$, Gender (female), and L1 Dutch concepts. The $\ge\!A2$ and Dutch concepts move further below the no-sensitivity baseline of $1$ from Layer 1 to Layer 2, while Gender stays close to $1$.}
    \label{fig:bert_bgr}
\end{figure}

$B_{\mathrm{gr}} < 1$ consistently for proficiency concepts. For BERT this moves further from $1$ with depth: $\ge\!A2$ drops from $0.81$ at Layer 1 to $0.46$ at Layer 2, with the same trend for $\ge\!B1$ and $\ge\!B2$. The BERT grader thus becomes more sensitive to proficiency closer to its output. The L1 directions reflect the dataset's score imbalance, with Dutch speakers scoring $0.82$ above non-Dutch on average and their $B_{\mathrm{gr}}$ dropping from $0.91$ to $0.52$ across the two BERT layers, while Gujarati speakers score $0.45$ below and its $B_{\mathrm{gr}}$ rises from $1.03$ to $1.10$, in the score-decreasing direction. The Whisper grader, by contrast, stays close to $B_{\mathrm{gr}} = 1$ for all demographic and L1 concepts and shows no clean depth trend. Also, its proficiency sensitivity is weaker than BERT's, reaching at most $0.83$ against BERT's $0.46$. Fig.~\ref{fig:bert_bgr} shows the per-speaker $B_{\mathrm{gr}}$ for BERT at both layers, making the deepening sensitivity to the $\ge\!A2$ and Dutch concepts visible, while the Gender concept stays close to the baseline at both layers.


These differences can be related to the input modality. BERT scores transcripts and can access speaker attributes only as far as they surface in lexical and grammatical choices, whereas Whisper has direct access to acoustic information such as pitch and accent that correlates with gender and L1. This is consistent with the recoverability results, where these concepts were linearly separable only at \texttt{dense.in}. 
The one within-model effect not explained by a score gap is a very slight sensitivity against the female gender concept at BERT Layer 2 ($B_{\mathrm{gr}} = 1.11$), despite the almost-zero gender score gap. These profiency and architecture trends are also found in S\&I, with BERT again becoming more sensitive with depth ($\ge\!B1$ reaching $0.23$ at Layer 2) and Whisper remaining near baseline. 

\begin{table*}[!ht]
\centering
\footnotesize
\setlength{\tabcolsep}{5pt}
\caption{Activation-space ($B_{\mathrm{gr}}$) and SAE-CAV ($B_{\mathrm{gr,SAE}}$) sensitivity on BULATS, mean $\pm$ standard deviation over five seeds. 
A star ($^*$) marks a weakly usable underlying CAV.}
\label{tab:bgr}
\begin{tabular}{l cccc cccc}
\toprule
& \multicolumn{4}{c}{Activation-space CAV} & \multicolumn{4}{c}{SAE-CAV} \\
\cmidrule(lr){2-5}\cmidrule(lr){6-9}
& \multicolumn{2}{c}{BERT} & \multicolumn{2}{c}{Whisper}
& \multicolumn{2}{c}{BERT} & \multicolumn{2}{c}{Whisper} \\
\cmidrule(lr){2-3}\cmidrule(lr){4-5}\cmidrule(lr){6-7}\cmidrule(lr){8-9}
Concept & Layer 1 & Layer 2 & dense.in & act.out & Layer 1 & Layer 2 & dense.in & act.out \\
\midrule
\multicolumn{9}{l}{\textit{Proficiency}} \\
$\ge\!A2$ & $0.81_{\pm.01}$ & $0.46_{\pm.03}$ & $0.83_{\pm.02}$ & $0.89_{\pm.00}$ & $0.82_{\pm.02}$ & $0.67_{\pm.05}$ & $0.95_{\pm.01}$ & $0.94_{\pm.03}$ \\
$\ge\!B1$ & $0.83_{\pm.01}$ & $0.63_{\pm.05}$ & $0.92_{\pm.01}$ & $0.88_{\pm.00}$ & $0.95_{\pm.02}$ & $0.71_{\pm.12}$ & $0.96_{\pm.01}$ & $1.01_{\pm.04}$ \\
$\ge\!B2$ & $0.89_{\pm.00}$ & $0.76_{\pm.05}$ & $0.93_{\pm.01}$ & $0.89_{\pm.00}$ & $0.94_{\pm.01}$ & $0.68_{\pm.07}$ & $0.95_{\pm.01}$ & $0.98_{\pm.03}$ \\
\addlinespace
\multicolumn{9}{l}{\textit{Demographic}} \\
Gender       & $0.99_{\pm.00}^*$ & $1.11_{\pm.02}^*$ & $1.02_{\pm.00}$ & $1.04_{\pm.00}$ & $0.99_{\pm.02}^*$ & $1.09_{\pm.12}^*$ & $1.00_{\pm.01}$ & $1.01_{\pm.08}$ \\
Age $\le\!30$ & $1.02_{\pm.00}$ & $1.06_{\pm.01}^*$ & $1.01_{\pm.00}$ & $1.02_{\pm.00}^*$ & $1.00_{\pm.02}$ & $1.06_{\pm.06}^*$ & $1.00_{\pm.01}$ & $0.99_{\pm.03}^*$ \\
\addlinespace
\multicolumn{9}{l}{\textit{First language (L1)}} \\
Dutch    & $0.91_{\pm.01}$ & $0.52_{\pm.04}$ & $0.96_{\pm.00}$ & $0.89_{\pm.00}$ & $0.98_{\pm.03}$ & $0.92_{\pm.04}$ & $0.96_{\pm.01}$ & $0.97_{\pm.07}$ \\
Gujarati & $1.03_{\pm.01}^*$ & $1.10_{\pm.01}$ & $1.00_{\pm.00}$ & $1.04_{\pm.00}$ & $1.02_{\pm.02}^*$ & $1.15_{\pm.10}$ & $1.01_{\pm.01}$ & $1.01_{\pm.03}$ \\
\bottomrule
\end{tabular}
\end{table*}

\begin{table}[!ht]
\centering
\small
\setlength{\tabcolsep}{4pt}
\caption{Selected TopK SAE configurations for BULATS: latent width $m$,
active units $k$, resulting sparsity, and dead-feature fraction.}
\label{tab:sae_configs}
\begin{tabular}{llrrrr}
\toprule
Model & Layer & $d$ & $m$ & $k$ & Sparsity \\
\midrule
BERT    & Layer 1  & 600  & 2400 & 320 & 0.867 \\
BERT    & Layer 2  & 20   & 80   & 16  & 0.800 \\
Whisper & dense.in & 2048 & 6144 & 480 & 0.922 \\
Whisper & act.out  & 64   & 256  & 24  & 0.906 \\
\bottomrule
\end{tabular}
\end{table}

\begin{figure}[!t]
\centering
\includegraphics[width=0.6\linewidth]{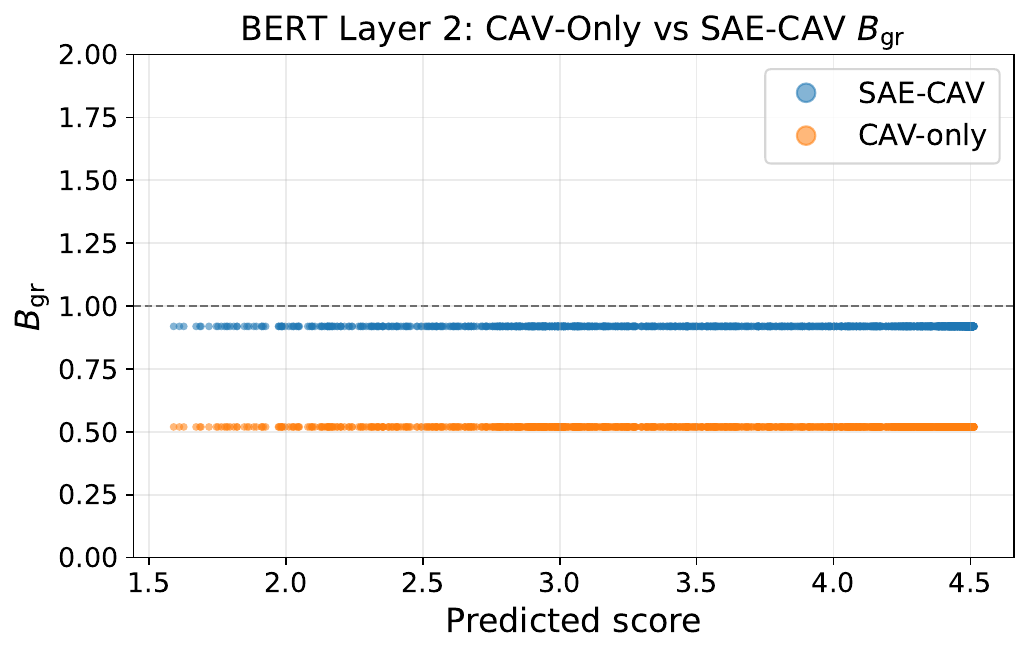}
\caption{Per-speaker $B_{\mathrm{gr}}$ for the L1 Dutch concept at BERT Layer 2, comparing the activation-space CAV with the SAE-CAV. The SAE-CAV is
attenuated towards the no-sensitivity baseline of $1$.}
\label{fig:sae_effect}
\end{figure}

\subsection{Bias Detection with Sparse Autoencoders}
\label{subsec:bias_sae}

We repeat the analysis in the SAE latent space, using the configurations in Table~\ref{tab:sae_configs}, selected per layer to balance reconstruction, sparsity, dead-feature fraction, and coherence. In latent space the concepts
become noticeably more recoverable. No concept remains degenerate, and several weakly usable directions become usable, confirming that the sparse space makes the concepts more linearly separable.

The SAE-CAV columns of Table~\ref{tab:bgr} show the influence results. The dominant pattern is that $B_{\mathrm{gr}}$ shifts towards the unbiased baseline of $1$, implying that a latent space probe that only partially recovers the concept direction measures weaker alignment with the score gradient. This
attenuation is strongest in the lowest-dimensional representation (BERT Layer 2, $20$-dimensional) and weakest in the highest (Whisper \texttt{dense.in}, $2048$-dimensional). The L1 Dutch concept spans both extremes, preserved almost exactly at \texttt{dense.in} ($0.96 \to 0.96$) yet
shifting by $+0.40$ at BERT Layer 2 ($0.52 \to 0.92$), shown per speaker in Fig.~\ref{fig:sae_effect}. The loss is therefore governed by the representation, not the concept. Because the SAE attenuates exactly the sensitivity signal that bias analysis depends on, and most severely in the small layers where bias was clearest, the sparse representation does not improve bias measurement over probing the activations directly.

\subsection{Results on S\&I}
\begin{table}[!t]
\centering
\small
\setlength{\tabcolsep}{5pt}
\caption{Activation-space $B_{\mathrm{gr}}$ on S\&I, mean $\pm$ standard deviation over five seeds. Conventions as in Table~\ref{tab:bgr}. A dash denotes a degenerate CAV, excluded from analysis.}
\label{tab:bgr_sandi}
\begin{tabular}{l cc cc}
\toprule
& \multicolumn{2}{c}{BERT} & \multicolumn{2}{c}{Whisper} \\
\cmidrule(lr){2-3}\cmidrule(lr){4-5}
Concept & Layer 1 & Layer 2 & dense.in & act.out \\
\midrule
$\ge\!B1$   & $0.77_{\pm.01}$ & $0.23_{\pm.01}$ & $0.77_{\pm.01}$ & $0.92_{\pm.00}$ \\
$\ge\!B2$   & $0.79_{\pm.00}$ & $0.73_{\pm.06}$ & $0.86_{\pm.01}$ & $0.92_{\pm.00}$ \\
Gender      & ---             & ---             & $1.02_{\pm.00}$ & $0.99_{\pm.00}^*$ \\
\bottomrule
\end{tabular}
\end{table}

Table \ref{tab:bgr_sandi} reports the activation space $B_{\mathrm{gr}}$ on S\&I, which reproduces the proficiency and architecture trends seen on BULATS. The BERT grader again becomes more sensitive with depth, with $\ge\!B1$ dropping from $0.77$ to $0.23$ from Layer 1 to 2. Whisper shows no depth trend and stays close to the baseline of 1. The CAV for the inferred gender concept is degenerate in both BERT layers for this dataset, and so its $B_{\mathrm{gr}}$ is omitted. It is recoverable at the Whisper head input, with $B_{\mathrm{gr}} = 1.02$, consistent with the absence of gender sensitivity on BULATS. 

\section{Conclusions}
\label{sec:conclusions}

We extended representation-level CAV bias analysis to two
modern speaking graders, a text-based BERT grader and a text-and-speech Whisper grader, and added a sparse autoencoder variant of the method. The central finding is that linear recoverability of a concept is governed by the representation it is measured in, not the concept itself, and that recoverability does not imply influence. This was seen in the Whisper grader, which encoded far more demographic information than BERT yet stayed close to the baseline of no sensitivity, while BERT grew more sensitive to proficiency with depth and
reflected the dataset's score imbalance through its L1 directions. For fairness analysis, this separation of encoding from influence is the key point, since an attribute being encoded is not in itself evidence of bias.

Applying the analysis in the sparse latent space did not improve it. The SAE made concepts more recoverable but attenuated the measured sensitivity towards the baseline, most severely in low-dimensional layers. The two metrics ask
different things. Recoverability only needs a classifier to separate the concept, whereas the gradient metric requires the decoder-mapped direction to still align with the score gradient in activation space, which the reconstruction objective has no reason to preserve. This motivates a score-aware SAE, trained to reconstruct the grader's output or to weight reconstruction error by the score gradient, so that the sparse space retains
the directions the bias metric needs.



\section*{Acknowledgments}

Generative AI tool, namely ChatGPT and Claude, was used only to review grammar and improve the fluency of the text.

\bibliographystyle{IEEEtran}
\bibliography{references}

@techreport{benedict2024thesis,
  author      = {Benedict Davies},
  title       = {{Automatic Assessment of English as a Second Language}},
  institution = {Department of Engineering, University of Cambridge},
  year        = {2024},
  type        = {{Fourth-Year Project Report}}
}

@inproceedings{kim2018tcav,
  title = {{Interpretability Beyond Feature Attribution: Quantitative Testing with Concept Activation Vectors (TCAV)}},
  author = {Kim, Been and Wattenberg, Martin and Gilmer, Justin and Cai, Carrie and Wexler, James and Vi{\'e}gas, Fernanda and Sayres, Rory},
  booktitle = {Proceedings of the 35th International Conference on Machine Learning},
  pages={2668--2677},
  year={2018},
  organization={PMLR}
}

@inproceedings{wei2021cavsla,
  title = {{Analysing Bias in Spoken Language Assessment Using Concept Activation Vectors}},
  author = {Wei, Xizi and Gales, Mark J. F. and Knill, Kate M.},
  booktitle = {{IEEE} International Conference on Acoustics, Speech and Signal Processing ({ICASSP})},
  pages={7753--7757},
  year={2021},
  organization={IEEE}
}

@article{bricken2023monosemanticity,
  title = {{Towards Monosemanticity: Decomposing Language Models With Dictionary Learning}},
  author = {Bricken, Trenton and Templeton, Adly and Batson, Joshua and Chen, Brian and Jermyn, Adam and Conerly, Tom and Turner, Nicholas L. and Anil, Cem and Denison, Carson and Askell, Amanda and Lasenby, Robert and Wu, Yifan and Kravec, Shauna and Schiefer, Nicholas and Maxwell, Tim and Joseph, Nicholas and Hatfield-Dodds, Zac and Tamkin, Alex and Nguyen, Karina and McLean, Brayden and Burke, Josiah E. and Hume, Tristan and Carter, Shan and Henighan, Tom and Olah, Christopher},
  journal = {Transformer Circuits Thread},
  year = {2023},
  url = {https://transformer-circuits.pub/2023/monosemantic-features/index.html}
}

@inproceedings{gao2024scaling,
  title = {{Scaling and Evaluating Sparse Autoencoders}},
  author = {Gao, Leo and Dupr{\'e} la Tour, Tom and Tillman, Henk and Goh, Gabriel and Troll, Rajan and Radford, Alec and Sutskever, Ilya and Leike, Jan and Wu, Jeffrey},
  booktitle = {International Conference on Learning Representations},
  volume={2025},
  pages={26721--26754},
  year={2025}
}

@inproceedings{radford2022whisper,
  title = {{Robust Speech Recognition via Large-Scale Weak Supervision}},
  author = {Radford, Alec and Kim, Jong Wook and Xu, Tao and Brockman, Greg and McLeavey, Christine and Sutskever, Ilya},
  booktitle = {Proceedings of the 40th International Conference on Machine Learning (ICML)},
  pages={28492--28518},
  year={2023},
  organization={PMLR}
}

@inproceedings{devlin2019bert,
  title = {{BERT: Pre-training of Deep Bidirectional Transformers for Language Understanding}},
  author = {Devlin, Jacob and Chang, Ming-Wei and Lee, Kenton and Toutanova, Kristina},
  booktitle = {Proceedings of the 2019 Conference of the North American Chapter of the Association for Computational Linguistics: Human Language Technologies, Volume 1 (Long and Short Papers)},
  pages={4171--4186},
  year = {2019},
}

@inproceedings{raina20_interspeech,
  title     = {{Universal Adversarial Attacks on Spoken Language Assessment Systems}},
  author    = {Vyas Raina and Mark J.F. Gales and Kate M. Knill},
  year      = {2020},
  booktitle = {{Interspeech 2020}},
  pages     = {3855--3859},
  doi       = {10.21437/Interspeech.2020-1890},
  issn      = {2958-1796},
}

@article{sicorpus25,
  author = {Kate Knill and Diane Nicholls and Mark J.F. Gales and Mengjie Qian and Pawel Stroinski},
  year = {2025},
  title = {{The Speak \& Improve Corpus 2025: an L2 English Speech Corpus for Language Assessment and Feedback}},
  publisher = {Cambridge University Press & Assessment},
  url = {https://doi.org/10.17863/CAM.114333}
}

@inproceedings{knill2025speakimprove,
  title = {{Introducing the Speak \& Improve Corpus 2025: an {L2} English Speech Corpus for Language Assessment and Feedback}},
  author = {Knill, Kate M. and Nicholls, Diane and Gales, Mark J. F. and Qian, Mengjie and Stroinski, Pawel},
  booktitle = {Proceedings of the 10th Workshop on Speech and Language Technology in Education ({SLaTE})},
  year = {2025},
  pages     = {167--171},
  doi       = {10.21437/SLaTE.2025-34},
  issn      = {2311-4975},
}

@misc{huh_voice_gender_classifier_github,
  author       = {Huh, Jaesung},
  title        = {{Voice Gender Classifier}},
  howpublished = {\url{https://github.com/JaesungHuh/voice-gender-classifier}},
  note         = {GitHub repository. Accessed November 2025}
}

@inproceedings{mcknight23_slate,
  title     = {{Automatic Assessment of Conversational Speaking Tests}},
  author    = {Simon W McKnight and Arda Civelekoglu and Mark Gales and Stefano Bann{\`o} and Adian Liusie and Katherine M Knill},
  year      = {2023},
  booktitle = {{9th Workshop on Speech and Language Technology in Education (SLaTE)}},
  pages     = {99--103},
  doi       = {10.21437/SLaTE.2023-19},
  issn      = {2311-4975},
}

@INPROCEEDINGS{banno2022slt,
  author={Bann{\`o}, Stefano and Matassoni, Marco},
  booktitle={2022 IEEE Spoken Language Technology Workshop (SLT)}, 
  title={{Proficiency Assessment of L2 Spoken English Using Wav2Vec 2.0}}, 
  year={2023},
  volume={},
  number={},
  pages={1088-1095},
  doi={10.1109/SLT54892.2023.10023019}}

@inproceedings{lin25_slate,
  title     = {{The NTNU System at the S\&I Challenge 2025 SLA Open Track}},
  author    = {Hong-Yun Lin and Tien Hong Lo and Yu Hsuan Fang and Jhen Ke Lin and Chung Chun Wang and Hao Chien Lu and Berlin Chen},
  year      = {2025},
  booktitle = {{10th Workshop on Speech and Language Technology in Education (SLaTE)}},
  pages     = {148--152},
  doi       = {10.21437/SLaTE.2025-30},
  issn      = {2311-4975},
}

@inproceedings{bolukbasi2016debiasing,
  title = {{Man is to Computer Programmer as Woman is to Homemaker? Debiasing Word Embeddings}},
  author = {Bolukbasi, Tolga and Chang, Kai-Wei and Zou, James and Saligrama, Venkatesh and Kalai, Adam},
  booktitle = {Advances in Neural Information Processing Systems 29 (NeurIPS)},
  year = {2016}
}

@inproceedings{buolamwini2018gendershades,
  title = {{Gender Shades: Intersectional Accuracy Disparities in Commercial Gender Classification}},
  author = {Buolamwini, Joy and Gebru, Timnit},
  booktitle = {Proceedings of the 1st Conference on Fairness, Accountability and Transparency (FAccT)},
  pages={77--91},
  year = {2018},
  organization={PMLR}
}

@inproceedings{bau2017networkdissection,
  title = {{Network Dissection: Quantifying Interpretability of Deep Visual Representations}},
  author = {Bau, David and Zhou, Bolei and Khosla, Aditya and Oliva, Aude and Torralba, Antonio},
  booktitle = {Proceedings of the IEEE conference on computer vision and pattern recognition (CVPR)},
  pages={6541--6549},
  year = {2017}
}

@inproceedings{alain2017probes,
  title = {{Understanding Intermediate Layers Using Linear Classifier Probes}},
  author = {Alain, Guillaume and Bengio, Yoshua},
  booktitle = {International Conference on Learning Representations (ICLR) Workshop},
  year = {2017}
}

@inproceedings{banno2023assessment,
  title={{Assessment of L2 Oral Proficiency Using Self-Supervised Speech Representation Learning}},
  author={Bann{\`o}, Stefano and Knill, Kate and Matassoni, Marco and Raina, Vyas and Gales, Mark},
  booktitle={Proceedings of 9th Workshop on Speech and Language Technology in Education (SLaTE)},
  pages={126--130},
  year={2023},
  doi       = {10.21437/SLaTE.2023-24},
  issn      = {2311-4975},
}

@inproceedings{ma2025b_interspeech,
  title     = {{Assessment of L2 Oral Proficiency using Speech Large Language Models}},
  author    = {Rao Ma and Mengjie Qian and Siyuan Tang and Stefano Bannò and Kate M. Knill and Mark J.F. Gales},
  year      = {2025},
  booktitle = {{Interspeech 2025}},
  pages     = {5078--5082},
  doi       = {10.21437/Interspeech.2025-1793},
  issn      = {2958-1796},
}

@inproceedings{baevski2020wav2vec2,
  title={{wav2vec 2.0: A framework for self-supervised learning of speech representations}},
  author={Baevski, Alexei and Zhou, Yuhao and Mohamed, Abdelrahman and Auli, Michael},
  journal={Advances in neural information processing systems},
  volume={33},
  pages={12449--12460},
  year={2020}
}

@inproceedings{desplanques20_interspeech,
  title     = {{ECAPA-TDNN: Emphasized Channel Attention, Propagation and Aggregation in TDNN Based Speaker Verification}},
  author    = {Brecht Desplanques and Jenthe Thienpondt and Kris Demuynck},
  year      = {2020},
  booktitle = {{Interspeech 2020}},
  pages     = {3830--3834},
  doi       = {10.21437/Interspeech.2020-2650},
  issn      = {2958-1796},
}

@inproceedings{phan25_slate,
  title     = {{One Whisper to Grade Them All}},
  author    = {Nhan Phan and Anusha Porwal and Yaroslav Getman and Ekaterina Voskoboinik and Tamás Grósz and Mikko Kurimo},
  year      = {2025},
  booktitle = {{10th Workshop on Speech and Language Technology in Education (SLaTE)}},
  pages     = {56--60},
  doi       = {10.21437/SLaTE.2025-12},
  issn      = {2311-4975},
}

@book{cefr2001,
author= {{Council of Europe}},
title={Common European Framework of Reference for Languages: Learning, Teaching, Assessment},
address={Cambridge},
publisher={Cambridge University Press},
year=2001,
url={https://rm.coe.int/1680459f97}
}

@article{chambers2011bulats,
  author  = {Lucy Chambers and Kate Ingham},
  title   = {{The BULATS Online Speaking Test}},
  journal = {Research Notes},
  year    = {2011},
  publisher = {University of Cambridge ESOL Examinations},
  url     = {https://www.cambridgeenglish.org/Images/23161-research-notes-43.pdf}
}

@inproceedings{qian25_slate,
  title     = {{Speak \& Improve Challenge 2025}},
  author    = {Mengjie Qian and Kate M. Knill and Stefano Bannò and Siyuan Tang and Penny Karanasou and Mark J.F. Gales and Diane Nicholls},
  year      = {2025},
  booktitle = {{10th Workshop on Speech and Language Technology in Education (SLaTE)}},
  pages     = {41--45},
  doi       = {10.21437/SLaTE.2025-9},
  issn      = {2311-4975},
}

@article{qian2024speak,
  title={{Speak \& improve challenge 2025: Tasks and baseline systems}},
  author={Qian, Mengjie and Knill, Kate and Banno, Stefano and Tang, Siyuan and Karanasou, Penny and Gales, Mark JF and Nicholls, Diane},
  journal={arXiv preprint arXiv:2412.11985},
  year={2024}
}


\end{document}